%% file: _main.tex
\documentclass[letterpaper]{article}
\usepackage{isea}
\usepackage[pdftex]{graphicx}
\usepackage{times}
\usepackage{helvet}
\usepackage{courier}

\usepackage{booktabs}       

\usepackage{subfig}
\usepackage{graphicx}

\usepackage{etoolbox}
\AtBeginEnvironment{quote}{\small}

\newlength\myparskip
\makeatletter
\renewenvironment{quote}%
  {\list{}{\rightmargin\leftmargin}%
  \parskip=\myparskip\item[]}%
  {\endlist}

\usepackage{url}

\usepackage{dblfloatfix}    
\usepackage{enumitem}

\usepackage{float}

\usepackage{kotex} 
 
\usepackage[numbers]{natbib}
\title{The Myths of Our Time: Fake News}

\author{
  Vít Růžička, Eunsu Kang, David Gordon, \\
  \Large\bf{Ankita Patel, Jacqui Fashimpaur, Manzil Zaheer} \\
  Carnegie Mellon University\\
  \texttt{previtus@gmail.com, eunsuk@andrew.cmu.edu, davidgor@andrew.cmu.edu,} \\
  \texttt{ankitap1@alumni.cmu.edu, jfashimp@andrew.cmu.edu, manzilz@andrew.cmu.edu} \\
}

\begin{document} 
\maketitle
\begin{abstract}
While the purpose of most fake news is misinformation and political propaganda, our team sees it as a new type of myth that is created by people in the age of internet identities and artificial intelligence. Seeking insights on the fear and desire hidden underneath these modified or generated stories, we use machine learning methods to generate fake articles and present them in the form of an online news blog. This paper aims to share the details of our pipeline and the techniques used for full generation of fake news, from dataset collection to presentation as a media art project on the internet.
\end{abstract}



\input{introduction}
\input{method_longer_from_4_page_version.tex}

\input{results}

\bibliographystyle{isea}
\bibliography{bibliography}

\input{supplementary}
\section{Authors Biographies}


\noindent \textbf{Vít Růžička} received his B.Sc. and M.Sc. with Honors in Computer Sciences with specialization in Machine Learning, Computer Graphics and Interaction from the Czech Technical University in Prague, Czech Republic in 2017. He spent nearly two exciting years of research internships at the Electrical and Computer Engineering department of Carnegie Mellon University in USA (September 2017 - May 2018) and at the EcoVision lab of the Photogrammetry and Remote Sensing group at ETH Zürich in Switzerland (January 2019 - July 2019). His research interest are Machine Learning, its application to other disciplines, Computer Vision, Creative AI and the intersections of Machine Learning and Art.
\smallskip

\noindent \textbf{Eunsu Kang} is a Korean media artist who creates interactive audiovisual installations and AI artworks. Her current research is focused on creative AI and artistic expressions generated by Machine Learning algorithms. Creating interdisciplinary projects, her signature has been seamless integration of art disciplines and innovative techniques. Her work has been invited to numerous places around the world including Korea, Japan, China, Switzerland, Sweden, France, Germany, and the US. All ten of her solo shows, consisting of individual or collaborative projects, were invited or awarded. She has won the Korean National Grant for Arts three times. Her researches have been presented at prestigious conferences including ACM, ICMC, ISEA, and NeurIPS. Kang earned her Ph.D. in Digital Arts and Experimental Media from DXARTS at the University of Washington. She received an MA in Media Arts and Technology from UCSB and an MFA from the Ewha Womans University. She had been a tenured art professor at the University of Akron for nine years and is currently a Visiting Professor with emphasis on Art and Machine Learning at the School of Computer Science, Carnegie Mellon University. 
\smallskip

\noindent \textbf{David Gordon} is an interdisciplinary artist and engineer living in the Los Angeles area. He has a specialty in simulation for autonomous systems, and currently works in NVIDIA's autonomous driving simulation division. He received a BCSA (Bachelors in Computer Science and Arts) from Carnegie Mellon University in 2019.
\smallskip

\noindent \textbf{Manzil Zaheer} earned his Ph.D. degree in Machine Learning from the School of Computer Science at Carnegie Mellon University under the able guidance of Prof Barnabas Poczos, Prof Ruslan Salakhutdinov, and Prof Alexander Smola. He is the winner of Oracle Fellowship in 2015. His research interests broadly lie in representation learning. He is interested in developing large-scale inference algorithms for representation learning, both discrete ones using graphical models and continuous with deep networks, for all kinds of data. He enjoy learning and implementing complicated statistical inference, data-parallelism, and algorithms in a simple way.

\end{document}

%% file: introduction.tex
\section{Introduction}



Is fake news a new type of myth that people are creating in the age of internet and artificial intelligence? K. Shu et al says fake news can have many definitions, and one narrow definition is ``a news article that is intentionally and verifiably false.'' \cite{fakenewsdef} While the purpose of fake news, according to K. Starbird, is disinformation and political propaganda \cite{ks-alternative-story-disinformation}, it often gives us some insights into people's hidden fears and desires the same way myths, folk tales, and urban legends do (an example in footnote\footnote{News story \cite{web-truck-story}, which is about a man who was run over by a parade car that was carrying dancing people at the Queer Culture Festival in Jeju city of Korea, spread quickly with a photo of a man under a truck. The actual event details came out eventually: the man had crawled under the car himself and he was mildly injured during his resistance against the policeman who was pulling him out for his own safety. This fake news did not describe the actual event but what the writer wanted to see to be able to frame people of the festival as a danger to their society. While it stimulated confusion as intended and strengthened the hate-cartel, it also vividly revealed their fear and worldview that had been hidden under their social masks.}).
We generate fake news using Machine Learning (ML) algorithms in an attempt to create new myths of our time and share them in the form of an online blog, \url{www.newsby.ml}\footnote{NewsBy.Ml blog url: \url{www.newsby.ml}}. Our project is not designed to lure people into a specific perspective but to make a visible statement on this phenomenon in the context of art, which offers multi-layered provocations and interpretations. We plan to further develop this project and contribute to fake news detection research by providing a labeled dataset.

There have been other fake text generation projects such as generative reviews by Y.Yao et al \cite{fakereviews} and generative Harry Potter books \cite{harry-potter-gen}. Our project, however, focuses on generating longer articles from a dataset of texts with inconsistent writing styles.

	\begin{figure}[!]
		\centering
		\frame{
		\includegraphics[width=0.46\textwidth]{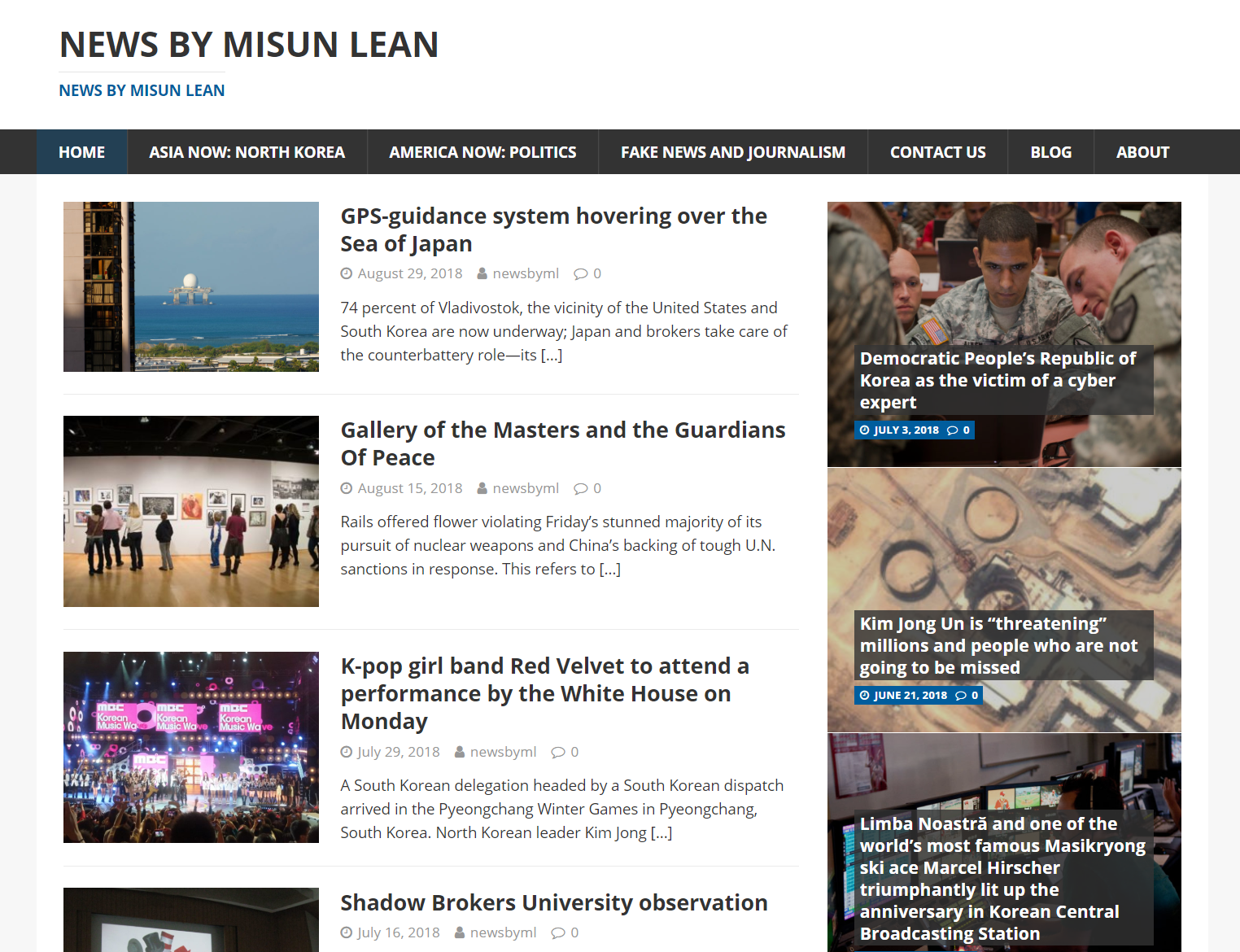}
		}
		\caption{Snapshot of the website used to present generated articles in a form of blog. Publicly available at \url{www.newsby.ml}.}
		\label{fig:webpage_screen}
	\end{figure}

%% file: method_longer_from_4_page_version.tex
\section{Method}

\begin{figure*}[!]
		\centering
		
		\includegraphics[width=0.95\textwidth]{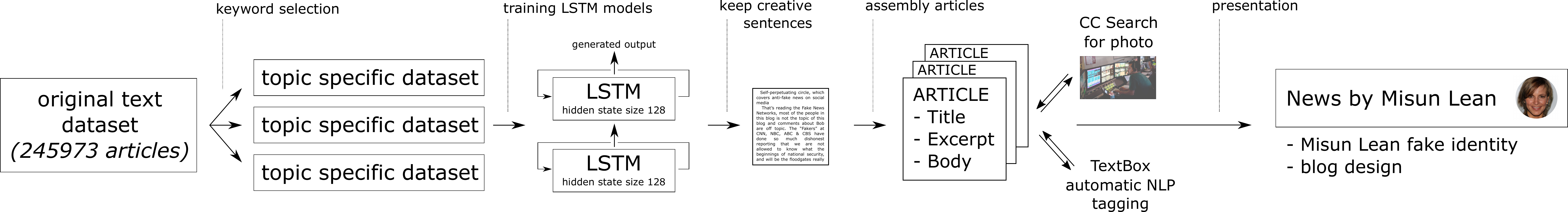}
		
		\caption{Fake news generation pipeline diagram}
		\label{fig:diagram}
\end{figure*}

The developed pipeline is presented in Figure \ref{fig:diagram}.

\subsection{Dataset collection and filtration}

\begin{table*}[]
\centering
\caption{Topics and keywords used to create specialized datasets.}
\label{topics_keywords_table}
\resizebox{\textwidth}{!}{%
\begin{tabular}{@{}llll@{}}
\toprule
topic & keywords & \#articles & \#words \\ \midrule
Asia Now: North Korea & \textit{'korea', 'korean', 'koreas', 'koreans', 'pyongyang', 'nkorea', 'jongun', 'jongil'} & 3064 & 13 572 577 \\
America Now: Politics & \textit{'trump', 'america', 'obama', 'obamas', 'american', 'americas', 'washington', 'california', 'fbi', 'mexico', 'florida'} & 9000 & 9 033 277 \\
Fake News and Journalism & \textit{\begin{tabular}[c]{@{}l@{}}'fake', 'truth', 'false', 'wrong', 'journalists', 'intelligence', 'zuckerberg', 'illegal', 'crime', 'terror', 'whistleblower', 'fbi', \\ 'cia', 'journalist', 'tweets', 'instagram', 'authorities', 'reporter', 'surveillance',  'allegations', 'wikileaks', 'controversial'\end{tabular}} & 9481 & 7 128 962 \\ \bottomrule
\end{tabular}

}
\end{table*}

In order to generate fake news, we need to collect a textual dataset corresponding to real world news articles. We started this effort by scraping news articles from websites. We used search terms from commonly used search APIs to get a large amount of generic articles, and also specifically searched with some topics in mind. This resulted in a large dataset of downloaded articles with varying topics. More specifically, we have assembled a dataset of 245,973 articles counting totally 196,952,689 words.

We then chose to select a few subsets of this initial large dataset to create several smaller filtered and biased datasets. Using a Natural Language Processing (NLP) tool from Machinebox Textbox \cite{web-machinebox}, we performed an entity extraction and automatic tag extraction on each article. We annotated our dataset with multiple tags describing the topics each article is addressing.

We manually created several categories of keywords, which are used to select subsets of the original large dataset. See these categories and their corresponding keywords in Table \ref{topics_keywords_table}. If an article is tagged with at least one of the labels from the selected keyword set, it will be added to the corresponding subset of articles.

This process creates unequally sized subsets of the original dataset, which can overlap with each other, but which also correspond to one selected topic.



\subsection{Training LSTM language models}

Having assembled several specialized datasets of articles, we train a language model for each one of the categories.

Language model tries to estimate the probability of sequence of words $w=(w_1,…,w_n)$ conditioned on the learned model as $p(w\mid model)$. Given the fact that each word depends on the previous words in the sequence, we use the chain rule of probability and estimate the conditional probability of the next word $w_i$ as:

\begin{equation}
  \prod_{i=1}^{n} p(w_i\mid w_1,…,w_{i-1},model)
\end{equation}


We are using the Long-Short Term Memory (LSTM) \cite{LSTM_paper, lstm_overview, rnn_language_model} Recurrent Neural Network to estimate $p(w_i\mid w_1,…,w_{i-1},model)$ while also using beam search \cite{sequence_transduction} when parsing the search tree. In this way we are creating specialized models, which are trained to learn the specific patterns in each of these subsets. Given the fact that each of these datasets contains a different number of articles and words in total, the model architectures can also be different. Larger datasets provide more data, and lend themselves to bigger models with more parameters. Generally, we use an LSTM model with two layers and 128 LSTM units.

The whole dataset of articles is converted into one large corpus of text, and is tokenized into vector representations of words. For the model we are using a github repository \textit{hunkim/word-rnn-tensorflow} implementation \cite{git-word-rnn-tensorflow}. We also experimented with converting each character into its vector representation (so called ``character based'' RNN), but empirically we received worse results with this method.

\subsection{Filtering generated text}

\begin{figure}[h]
		\centering
		\frame{
		\includegraphics[width=0.45\textwidth]{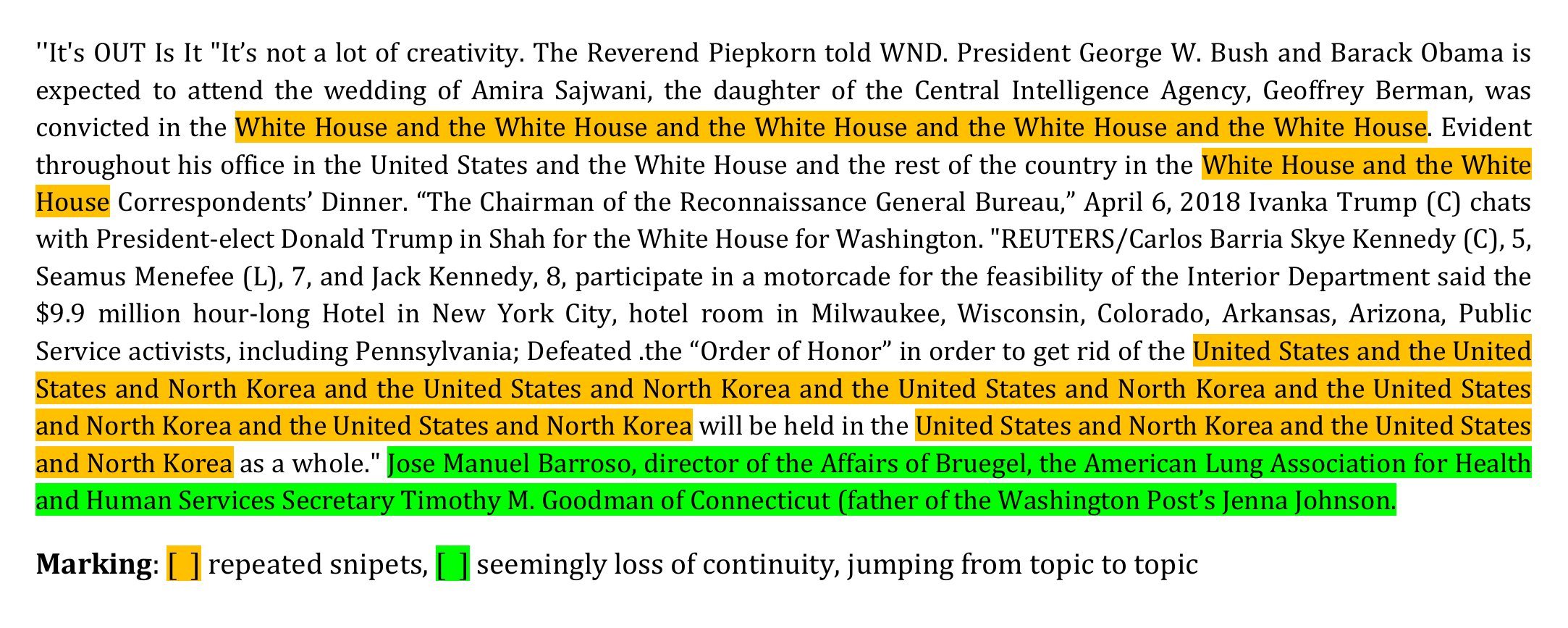}
		}
		\caption{Sample of generated raw text with problems that require manual corrections highlighted. We can see repeated fragments of text and topical loss of focus.}
		\label{fig:highlighted_problems}
	\end{figure}

Once LSTM models are trained to capture the underlying patterns of each individual dataset, we can generate a large amount of text from each of these models. The resulting generated text, however, contains samples which seem to be stuck in a loop or which mimic the original dataset perfectly without any innovation. See Figure \ref{fig:highlighted_problems} for an illustration of these problems.
We have chosen to analyze the generated text's novelty as compared with its original text dataset by comparing these two sets sentence by sentence. For each generated sentence we used the Levenstein distance with every sentence of the original dataset to obtain the closest match. We have decided to keep only the sentences which are more than $30\%$ dissimilar to their closest match in the original dataset. This gives us a large amount of generated and creative samples to choose from for the task of assembling the final fake news articles.

\input{results_generated_articles.tex}

\subsection{Article assembly}

Note that these selected and filtered creative generated text samples still contain a lot of sentences which would not fool a human judge. For example, there are sentences which have many repetitions, some syntactic and grammatical errors, and miss-spellings and also nonsense sentences. However, we discovered that these sentences are good enough to fool some fake news classifiers such as FakerFact \cite{web-fakerfact} or Textbox \cite{web-machinebox}. This shows how hard the problem of fake news detection is.

\begin{figure}[h]
		\centering
		
		\frame{
		\includegraphics[width=0.45\textwidth]{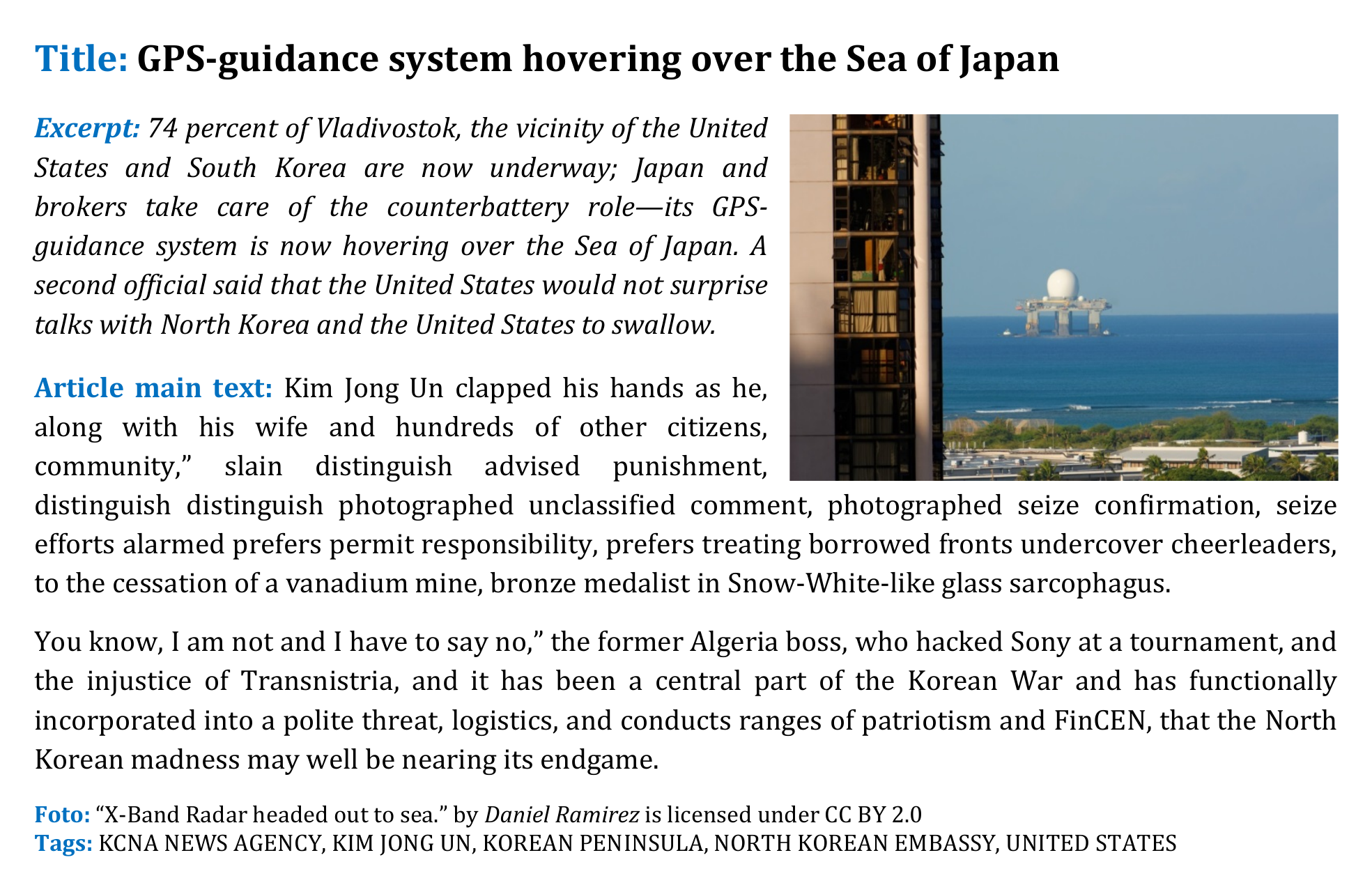}
		}

        \smallskip
		
		\frame{
		\includegraphics[width=0.45\textwidth]{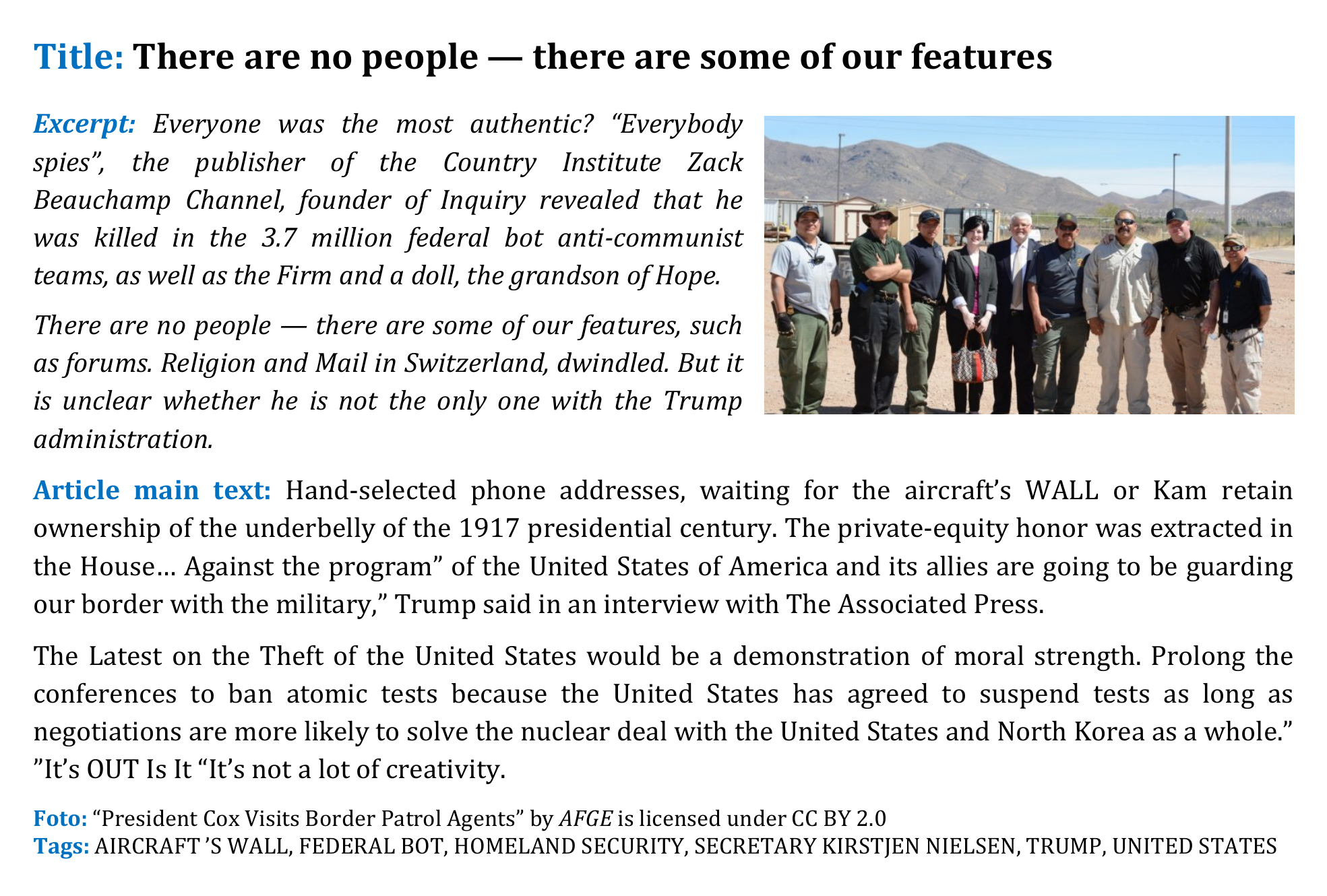}
		}
		\caption{Excerpts from the publicly available generated fake news articles, showing the article structure. Note that we show only a part of the article and that the tags were automatically generated by \cite{web-machinebox}.}
		\label{fig:sample_2}
	\end{figure}

We are assembling a new news article from these generated sentences for which we chose the following structure: (i) title, (ii) short excerpt used as a short description of the article and finally (iii) the main body of the article. See example articles in Figure \ref{fig:sample_2}.

We designed and used the below rules for a very limited amount of intervention when assembling article excerpts. 

\begin{itemize}[leftmargin=+.45in,rightmargin=+.4in]
    \item Around 50-100 words for the excerpts of each article. 
    \item We can delete or replace one letter from a word or one word from a sentence to fix obviously redundant elements.
    \item We can delete sentences that do not work with the overall subject of the paragraph.
    \item We can change the order of sentences to make the article more legible. 
\end{itemize}

For the longer article bodies, we further limited the amount of interventions, only allowing ourselves to throw away some completely broken sentences. 

Article titles were chosen by the authors from the article body. The illustrative image was selected by searching a database of Creative Commons images \cite{web-creativecommonssearch} using our article titles. We quote the author and the original name of the used photograph as required by the Creative Commons.

In both cases, the selection process was chosen to combat the current shortcomings of generative LSTM models with the least amount of intervention possible. 

\if false
\begin{table}[]
\centering
\caption{Rules used when assembling articles from the generated text.}
\label{rules_of_editing}
\begin{tabular}{@{}ll@{}}
\toprule
article & rules \\ \midrule
title & Manual selection, editing and capitalization \\
excerpt & Manual selection, reordering, fixing grammatical errors \\
body & Manual selection of full sentences, reordering \\ \bottomrule
\end{tabular}
\end{table}
\fi

\subsection{Tag generation}

After generating a number of articles for each thematic category, we present them on a newly built website. Finally we use an automated NLP tagging solution from Machinebox Textbox \cite{web-machinebox} to get realistic tags for each of these articles.


%% file: results_generated_articles.tex
\smallskip
\smallskip
\smallskip
\smallskip

\textbf{Some generated paragraph examples are:}

\smallskip
\smallskip

\begin{quote}
``A climate scientist at the RAND Corporation, said that the United States would prematurely withdraw from Syria,'' Trump wrote in The Washington Post. ... Trump scoffed. He also said he would not be able to comment on the notion. It is unclear whether he was not reimbursed by the White House and the Department of Homeland Security. ... ``Don’t worry, we are going to be able to get rid of the United States,'' he writes. ``I think that we are going to be great,'' Trump said in a statement that he was fired.''

(Category: \textit{America Now: Politics})
\end{quote}

\begin{quote}
``Seventeen-year-old Kim Jong Il was initially sentenced to 24 years in prison and fined 18 billion won (US\$16.8 million) on Twitter. He also took a group photo with them from Tiananmen Square, according to South Korean pop stars visiting Pyongyang.''

``Packs of wolves are coming from North Korea on Saturday to the United States in exchange for a freeze on North Korean exports of coal, iron, iron ore and seafood, oil and gas pipelines.''

(Category: \textit{Asia Now: North Korea})
\end{quote}

\begin{quote}
``European Federation of Journalists, today condemned the importance of the digitalization of mass media. Sarah Huckabee Sanders lambastes fake news on Russia Fraud Human Rights Watch and Society of Journalists.''

``Comments about Bob are off topic. The ``Fakers'' at CNN, NBC, ABC, CBS have done so much dishonest reporting that we are not allowed to know what the beginnings of national security.''

``We have a self-perpetuating circle, who covers anti-fake news on social media and media freedom in Croatia.''

(Category: \textit{Fake News and Journalism})
\end{quote}


\smallskip

%% file: results.tex
\smallskip

\section{Results}

\subsection{Fake news blog as an art project}

\begin{figure}[h]
		\centering
		\includegraphics[width=0.25\textwidth]{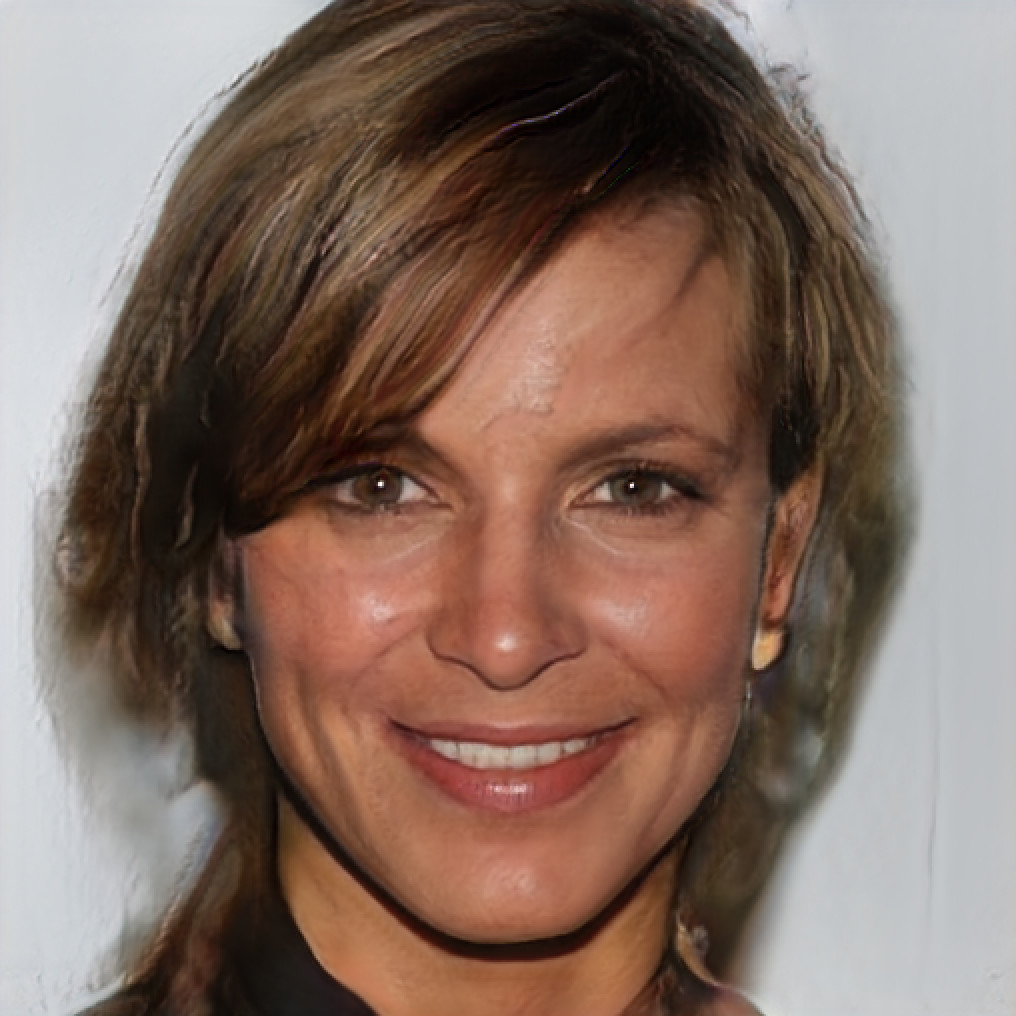}
		\caption{Portrait of a fake journalist Misun Lean generated by PGGAN \cite{pggan_paper}.}
		\label{fig-misun-lean}
	\end{figure}

Our project is presented in the form of a blog, which is one of the ways fake news is distributed online. The correspondent, Misun Lean, is a fake journalist identity created just for this project. Her name comes as a word play on the abbreviation ``ML'' (hence the blog name \textit{``News by ML''}) and her photograph shown in Figure \ref{fig-misun-lean} was generated using PGGAN \cite{pggan_paper}, a ML algorithm for generating high resolution portraits. 

We select article topics often used for spreading fake news, with focus on popular global issues and regions of conflict. In particular, we have selected \textit{Asia now: North Korea} and \textit{America now: Politics}. These topics are used either to focus the attention of readership on issues occurring in other countries, or to sway the opinions of masses during elections \cite{web-ny-fake-news-partisan, web-youtube-fakenews}. We also selected the topic of \textit{Fake news and Journalism} to serve as a mirror on how the problem of fake news is being covered by the media itself.


We designed the project website, shown in Figure \ref{fig:webpage_screen}, to follow the general structure of many news and opinion blogs. This website as a media art project proactively employs the culture and technology of current web environment to make an artistic statement on the phenomenon of fake news that spreads through internet. 

\subsection{Article credibility analysis}

We ran FakerFact \cite{web-fakerfact}, which is trained to classify the intention of online texts, on our generated news articles. The analysis did not detect any red flags, however some articles were classified as opinionated, sensational, or agenda driven. For example, the analysis said \textit{``Walt says it sounds the author may be more focused on pushing an agenda than sharing the facts''} for our article ``Israel continued to enlarge the ransacking of artists'', which was classified as Agenda Driven and \textit{``Hmm, Walt doesn't see strong red flags, but it's possible the author has a slight opinion''} for our article ``GPS-guidance system hovering over the Sea of Japan'', which was classified as Low Opinion. We consider this analysis to be consistent with our intention to appear as an opinionated political blog.






\if false
\begin{figure}%
    \centering
    \subfloat[America Now: Politics]{{\includegraphics[width=6cm]{figures/out_wordcloud_america10PLUS_allWords.png} }}%
    \qquad
    \subfloat[Fake News and Journalism]{{\includegraphics[width=6cm]{figures/out_wordcloud_fakenews10PLUS_allWords.png} }}%
    \caption{Word cloud visualization of the raw generated texts for different datasets and models, using only words which appeared with frequency of 10 and higher.}%
    \label{fig:word_clouds}%
\end{figure}
\fi

\if false
	\begin{figure}[t]
		\centering
		\frame{
		\includegraphics[width=0.8\textwidth]{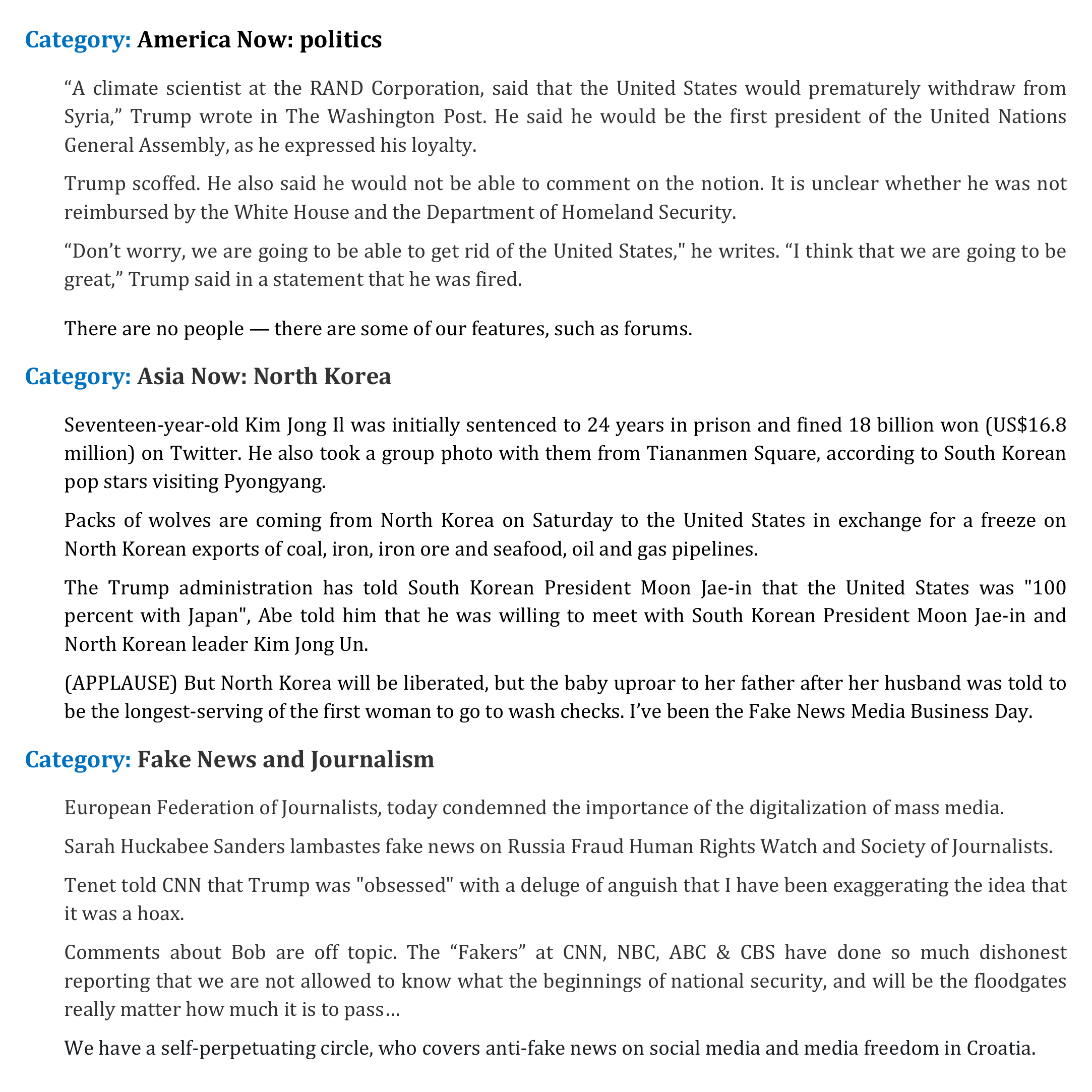}
		}
		\caption{Selected generated paragraph examples from each category.}
		\label{fig:selected-paragraphs}
	\end{figure}
\fi

\smallskip

\section{Conclusion}


For this project, we have created our own dataset, generated every integral part of a fake news blog including the correspondent's identity, and presented it in the form of online blog which acts as an art project to provoke conversations about fake news and the human desires behind this phenomenon. In this paper, we have shared the overall pipeline and details of our methods with a hope of helping other artists create more projects that discusses the phenomenon of fake news in our society as well as giving the general audience an understanding of the process of fake news generation. We share the project at GitHub \url{https://github.com/previtus/fake_news_generation_mark_I}.

For our future work, we plan to generate a new dataset category about Artificial Intelligence. We want to capture how it is being presented online, since a lot of excitement and fear surrounding it is based on false or exaggerated information. We are also considering creating a new dataset from ``alternative media'' websites that contribute to the propagation of fake news \cite{ks-alternative-story-disinformation}.